**Self-attention vector output similarities reveal how machines pay attention**

**Tal Halevi[a,1], Yarden Tzach[a,1], Ronit D. Gross[a], Shalom Rosner[a], and Ido Kanter[a,b,*]**

[a]Department of Physics, Bar-Ilan University, Ramat-Gan, 52900, Israel.

[b] Gonda Interdisciplinary Brain Research Center, Bar-Ilan University, Ramat-Gan, 52900, Israel.

*Corresponding author at: Department of Physics, Bar-Ilan University, Ramat-Gan, 52900, Israel.
E-mail address: ido.kanter@biu.ac.il (I. Kanter).

[1]These authors equally contributed to this work

## Abstract

The self-attention mechanism has significantly advanced the field of natural language processing, facilitating the development of advanced language-learning machines. Although its utility is widely acknowledged, the precise mechanisms of self-attention underlying its advanced learning and the quantitative characterization of this learning process remains an open research question. This study introduces a new approach for quantifying information processing within the self-attention mechanism. The analysis conducted on the BERT-12 architecture reveals that, in the final layers, the attention map focuses on sentence separator tokens, suggesting a practical approach to text segmentation based on semantic features. Based on the vector space emerging from the self-attention heads, a context similarity matrix, measuring the scalar product between two token vectors was derived, revealing distinct context similarities between different token vector pairs within each head and layer. The findings demonstrated that different attention heads within an attention block focused on different linguistic characteristics, such as identifying token repetitions in a given text or recognizing a token of common appearance in the text and its surrounding context. This specialization is also reflected in the distribution of distances between token vectors with high context similarities as the architecture progresses. The initial attention layers exhibit substantially long-range context similarities; however, as the layers progress, a more short-range context similarity develops, culminating in a preference for attention heads to create strong context similarities within the same sentence. Finally, the behavior of individual heads was analyzed by examining the uniqueness of their most common tokens in their high context similarity elements. Each head tends to focus on a unique token from the text and builds similarity pairs centered around it. This methodology for quantifying the behavior of the vector space emerging from the self-attention layers offers a novel way to understand the dynamics of the self-attention mechanism with respect to its outputs.

## 1. Introduction

Natural language processing (NLP) has been the core goal of researchers and developers since the advent of machines capable of learning [1-5]. The attention mechanism [5, 6] has significantly accelerated progress in NLP at an unprecedented rate. The core idea behind the attention mechanism, specifically the multi-head self-attention mechanism used in single-language comprehension, is to assess the importance of each component in the text relative to other elements. This is achieved by computing the context similarity between the key and query values of each token [7, 8] and producing new post-attention vectors by multiplying the attention map of each token by the token's value (Fig. 1). Although the model performs the relatively simple action of linear multiplication, the underlying processes within the attention mechanism remain a subject of ongoing investigation. A critical question is: can a quantifiable language explain the overall mechanism that makes this architecture effective for language learning? Some theories attempt to explain the attention mechanism by identifying specific semantic correlations or relations in the attention heads [9], or by modeling the system as a multidimensional spin model [10]. However, these approaches do not provide a quantitative account of the internal mechanism governing this architecture. This is because the calculation process does not end with the attention map but rather with the output vectors from each attention head, which then propagate to the output layer. Furthermore, greater focus is placed on understanding the attention map alone [11], neglecting a broader exploration of the output of each attention layer in its entirety, or the output of individual attention heads and the resulting vector space emerging from it. Notably, prior studies have demonstrated that the attention weights alone are not sufficient for explaining  the learning processes within this architecture [12]. Although the limitations of the attention map are apparent, a method for quantifying and measuring interactions within the emerging vector space of the attention head is required to further the understanding of the self-attention's mechanism.

This study investigates the behavior of the attention head outputs and their role in facilitating learning for NLP tasks. To better observe the vector space emerging from each attention head along the layers, the context similarity between each pair of output vectors was measured. This was achieved by computing the dot product between the unnormalized vector representations of each pair of the 128 tokens produced by the attention head. Unlike the attention weights in the attention map, this places greater emphasis on the input alterations of the token vector space. This creates a $128 \times 128$ matrix of context similarity scores, where higher values correspond to more strongly similar token representations, and lower values indicate weaker similarity. This interplay between alignment of the vectors and vector sizes offers insights into the behavior of the vectors emerging from attention, as well as which vectors are most influential in the signal propagating through the transformer architecture. Exploring the nature of the vector space and the cooperation between attention heads can yield a deeper understanding of how transformer architectures process language. The results were analyzed using the BERT-12 architecture [13]. The results display a new method of quantifying the learning process and shows how that learning progresses with the layers as well as across attention heads within each layer.

## 2. The Attention Mechanism and Context Similarity Matrix

NLP architectures process text using the following steps. Each sentence is converted into tokens, which are the basic building blocks of a sentence, typically representing a word or part of a word. Each token is embedded [4, 14, 15] into a 768-dimensional vector using an embedding layer. The embedding layer encodes both the lexical meaning of a token and its position within the input. The input sequence's length remains fixed, typically at 512 or 128 tokens, and any unused space is padded using a [PAD] token [11]. In this research a constant sequence length of 128 was used.

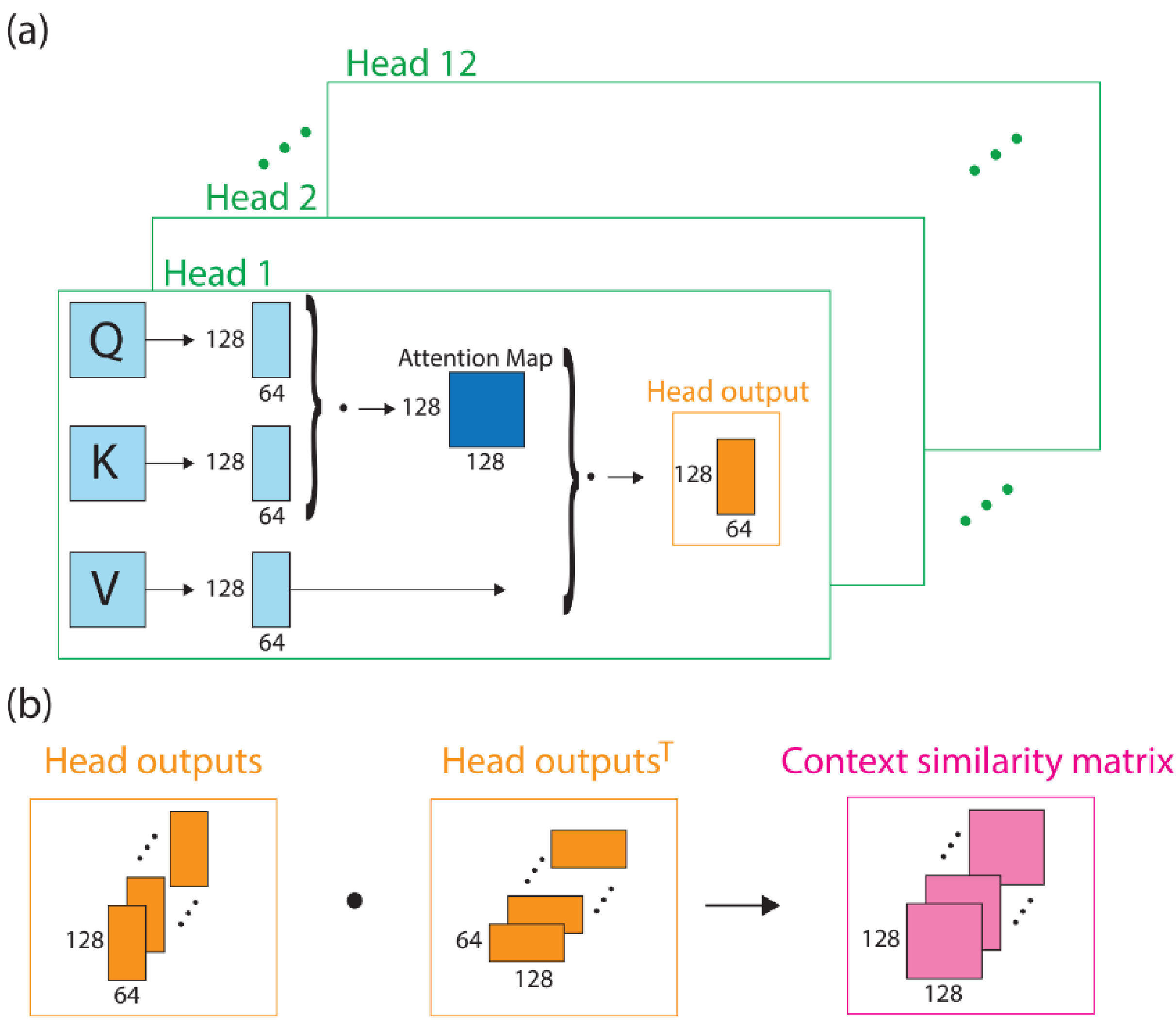

**Fig. 1.** Process of calculating attention map (a) and generating context similarity matrices (b) per individual attention head.

The input is now of shape $SequenceLength \times 768$ $(SequenceLength = 128)$ and is subsequently propagated through the transformer architecture, wherein at each attention layer, it is updated based on its attention weights into a vector space, enabling more effective learning. This raises the question of how these vectors evolve through the attention layers and how these changes relate to the representation of individual tokens and the overall meaning of a sequence.

The attention map, or attention weights, is a matrix of shape $Sequence\ Length \times Sequence\ Length$ showing the relative strength of each token's query multiplied by its key value [5]. The core concept is that strong attention weights indicate a strong relationship between two tokens. As such, the attention map shows how the architecture focuses on different regions of the text as the layers progress. These explanations provide a rudimentary understanding of the text's behavior and, unfortunately, do not satisfy the need for a quantifiable language capable of measuring or explaining how well each layer learns or what it learns. Attempting a macroscopic measure by summing over the attention maps of different text inputs does not reveal any fundamental features relevant to learning. It only shows that each head's summed attention weights are centered at the diagonal, with some presenting their strongest attention off-diagonal of either a single increment or decrement of 1 from the main diagonal. This shows that some heads tend to place more focus on the token itself, the subsequent token, or the preceding token (Fig. 2).

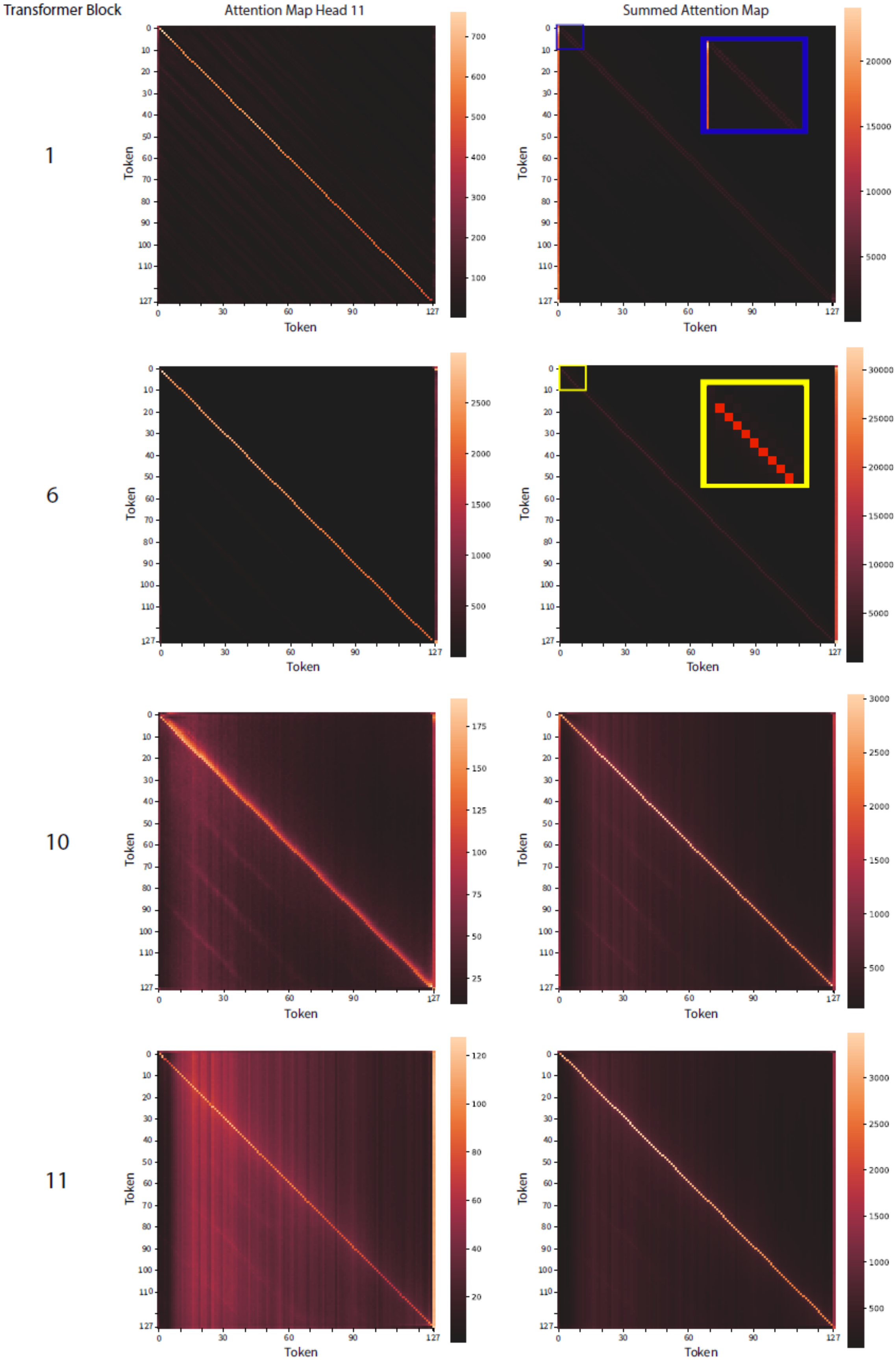

**Fig. 2.** Summed attention maps of an individual head (left) and summation of the attention maps of the 12 heads of a given layer over many inputs (right) for selected layers.

This diagonality is attributed to the weak diagonal elements that exist within the attention weights of the individual samples and are then accumulated together (Fig. 3). This reveals an ever-present diagonal attention procedure and might hint at a more diagonalized form of average attention focus; however, it significantly differs from the behavior of the attention maps of a single text input (Fig. 4).

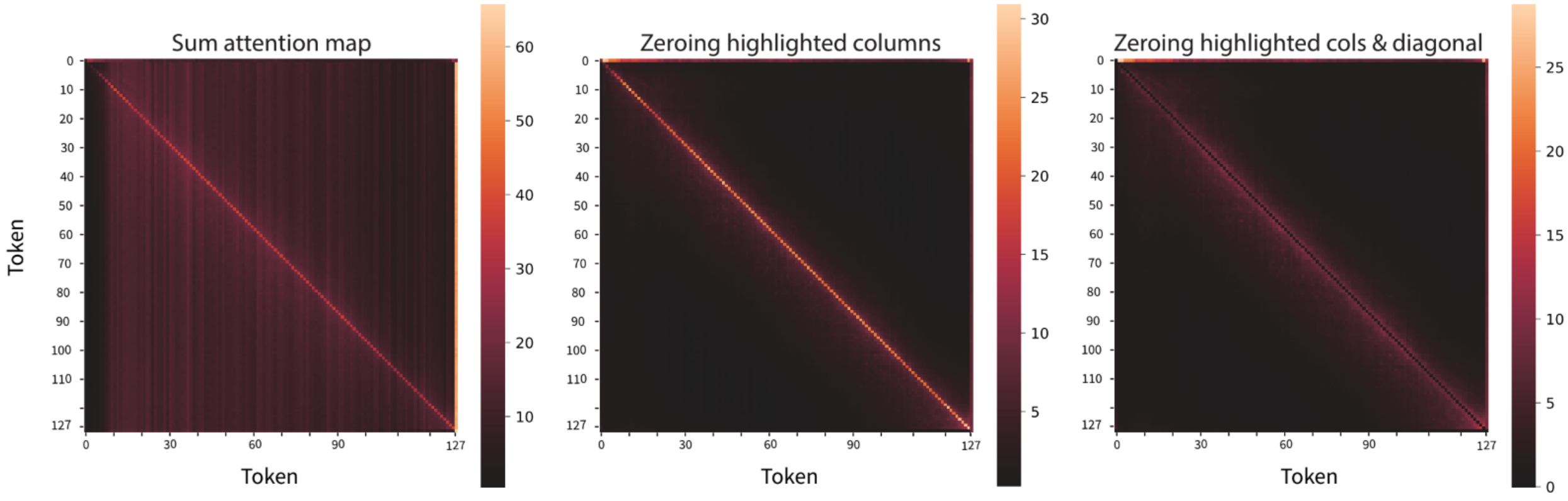

**Fig. 3.** Summed attention map of layer 11 (last layer) on a multitude of examples (left), summation with zeroing the most influential attention columns (middle), and summation with zeroing attention columns and diagonals (right).

Analyzing the behavior of the attention mechanism and attention head outputs reveals the following phenomena: For each input, the attention map undergoes a similar trend of centralization along the layers all the way to focusing attention on specified sentence separation tokens (“.”, “,”) (Fig. 4). At the initial layers the attention is more dispersed over the map of each head, and summing the attention reveals that the most common token for attention is “[CLS]” (a special token denoting the beginning of an input sequence) (Fig. 2, layer 1). In the middle layers (Fig. 4, layer 6), attention is more centralized along the diagonal, and in the last layers, attention is focused on sentence separators, a trend observed in many architectures [9].

After self-attention, the context similarity matrix is extrapolated to the attention head's output. Each attention head output of size

$$Sequence\ Length \times (\frac{Embedding\ dim}{\#\ heads})$$

is multiplied by its transpose to create a context similarity matrix (Fig. 1)

$$Context\ Similarity\ Matrix = head_{Output} \cdot head_{Output}^{T}$$

where in the presented results $SequenceLength = 128$.

This revealed areas of greater alignment and vector size between the output vectors of the attention mechanism. The diagonal was zeroed to indicate areas with high off-diagonal context similarity (Fig. 4) enabling the extrapolation of context similarities between different token vector representations. To better understand the context similarity matrix, the similarities of the vector space in the last layer are procured. In the last layer, the attention map for individual input samples forms column like attention on sentence separators (Fig. 4). This enables to use the attention map and group together areas of high context similarities based on high attention sentence separators. The attention map columns are summed, where the column sums (tokens) are normalized by their maximal value and clipped by a threshold to distinguish columns of high attention. These high attention token columns are then marked on both the X and Y axes of the context similarity matrix (Fig. 5). Thus, tokens with high attention (sentence separators) were used to delineate areas of common similarity in the context similarity matrix. The high-attention token columns and rows of the context similarity matrix were marked to show the delineation. These revealed areas of high-context similarity, in contrast to the low- context similarity in the head output vectors (Fig. 5, bottom right).

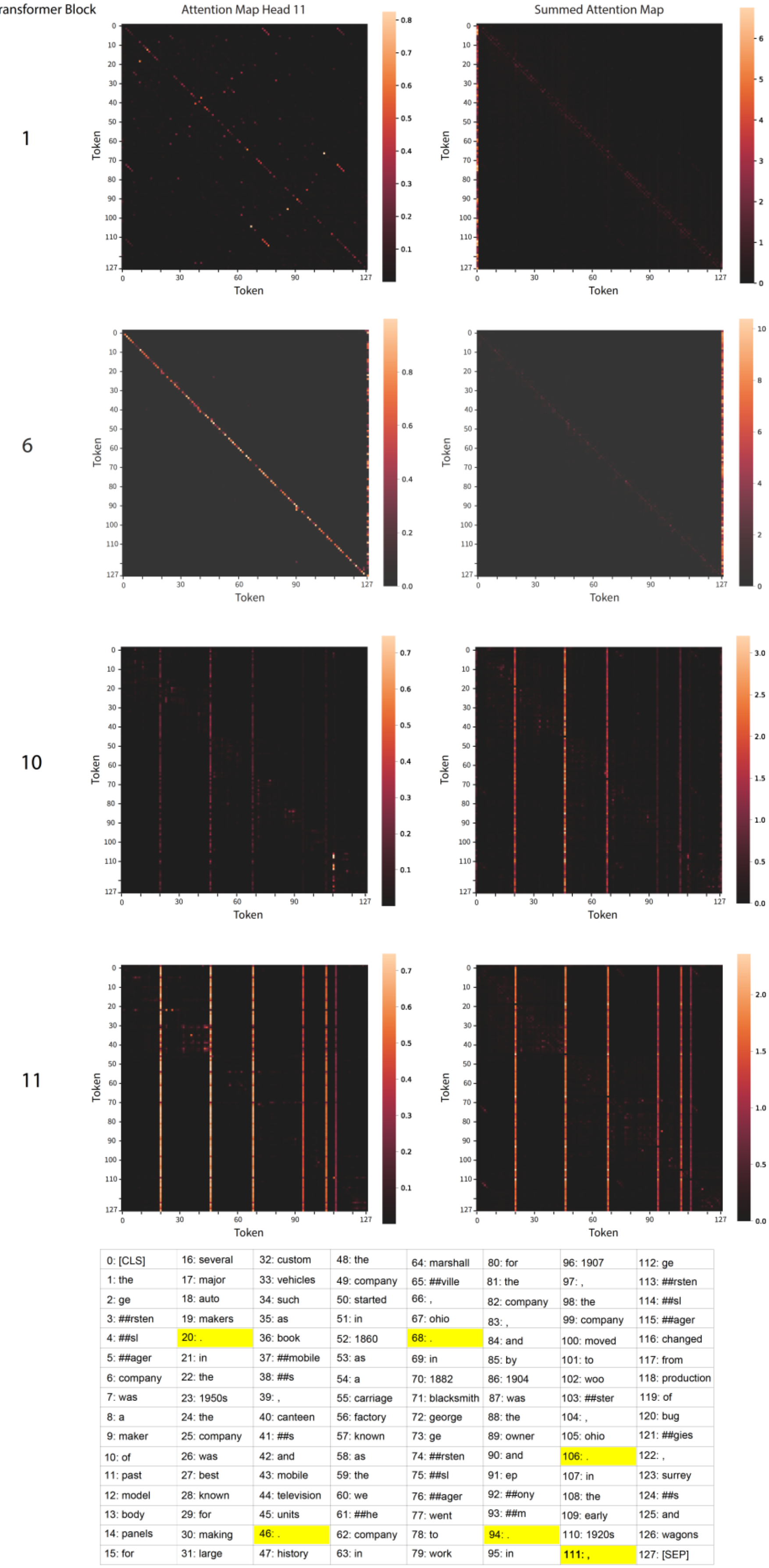

| 0: [CLS] | 16: several | 32: custom | 48: the | 64: marshall | 80: for | 96: 1907 | 112: ge |
|---|---|---|---|---|---|---|---|
| 1: the | 17: major | 33: vehicles | 49: company | 65: ##ville | 81: the | 97: , | 113: ##rsten |
| 2: ge | 18: auto | 34: such | 50: started | 66: , | 82: company | 98: the | 114: ##sl |
| 3: ##rsten | 19: makers | 35: as | 51: in | 67: ohio | 83: , | 99: company | 115: ##ager |
| 4: ##sl | 20: . | 36: book | 52: 1860 | 68: . | 84: and | 100: moved | 116: changed |
| 5: ##ager | 21: in | 37: ##mobile | 53: as | 69: in | 85: by | 101: to | 117: from |
| 6: company | 22: the | 38: ##s | 54: a | 70: 1882 | 86: 1904 | 102: woo | 118: production |
| 7: was | 23: 1950s | 39: , | 55: carriage | 71: blacksmith | 87: was | 103: ##ster | 119: of |
| 8: a | 24: the | 40: canteen | 56: factory | 72: george | 88: the | 104: , | 120: bug |
| 9: maker | 25: company | 41: ##s | 57: known | 73: ge | 89: owner | 105: ohio | 121: ##gies |
| 10: of | 26: was | 42: and | 58: as | 74: ##rsten | 90: and | 106: . | 122: , |
| 11: past | 27: best | 43: mobile | 59: the | 75: ##sl | 91: ep | 107: in | 123: surrey |
| 12: model | 28: known | 44: television | 60: we | 76: ##ager | 92: ##ony | 108: the | 124: ##s |
| 13: body | 29: for | 45: units | 61: ##he | 77: went | 93: ##m | 109: early | 125: and |
| 14: panels | 30: making | 46: . | 62: company | 78: to | 94: . | 110: 1920s | 126: wagons |
| 15: for | 31: large | 47: history | 63: in | 79: work | 95: in | 111: , | 127: [SEP] |

**Fig. 4.** Attention map of an individual head (left) and summation of all the attention maps per head of a given layer over a given input (right) for selected layers. The input tokens with marked areas of high attention (bottom).

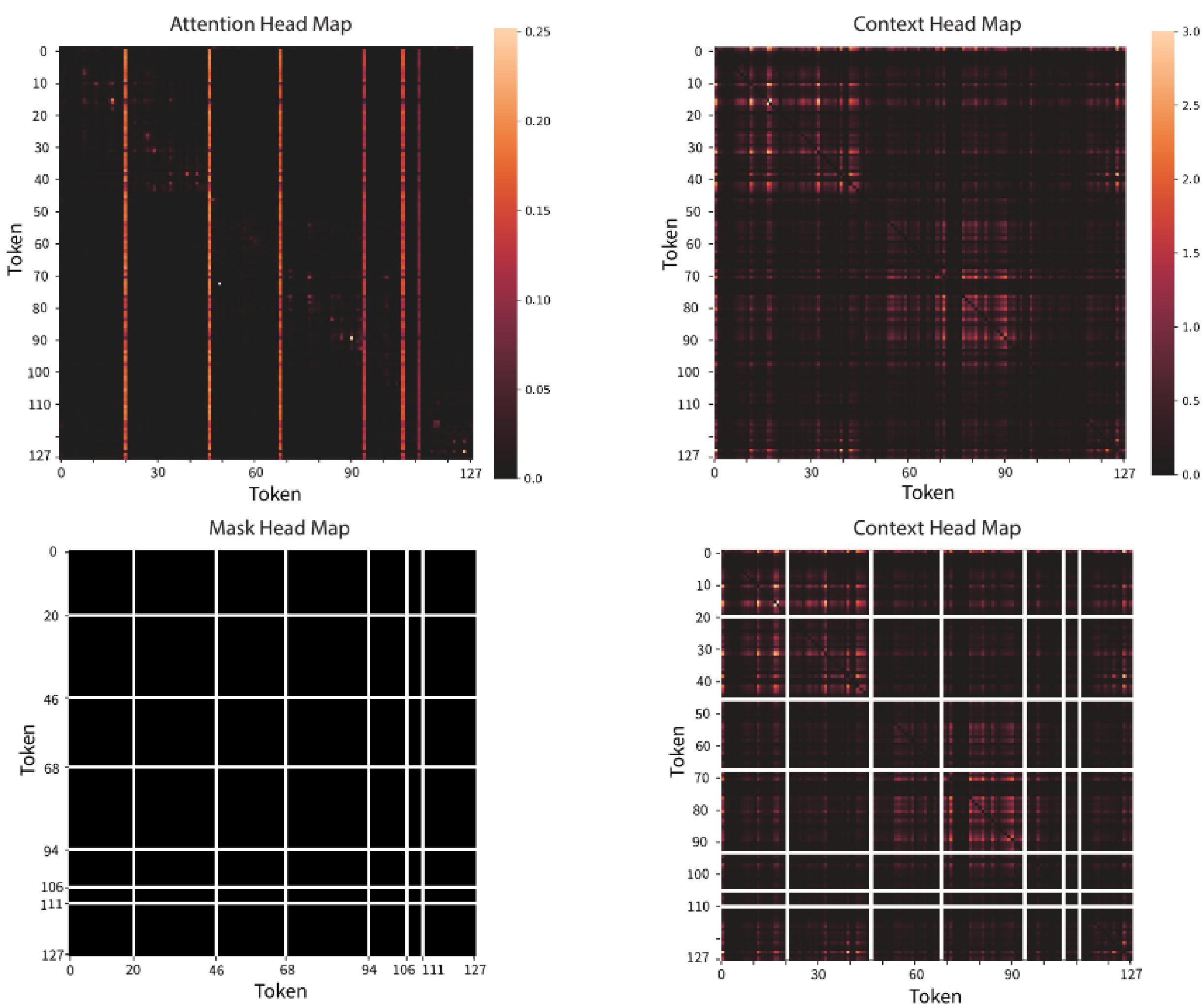

**Fig. 5.** Attention map of an individual head over a given input (top-left), context similarity matrix of head's output (top-right), masked head map based on attention lines and attention lines transposed (bottom-left), averaged regions of the context similarity matrix based on the attention masks (bottom-right).

Nearly all heads of the last layers exhibit similar attention maps in a columnar fashion. However, the difference lies in the distribution of the maximal-square elements. Each head had an area with many points of high-context similarity in different attention-confined squares, most commonly along the diagonal.

3. **Head Behavior:**

Analysis of the context similarity matrix enables quantification of the vector space emerging from each attention head. Two vectors with higher context similarity indicate a similar field output in the layer's head, or that the specific head amplifies the similarity or connection between these two vectors.

To better understand the learning behavior of different heads, the most significant attention map elements were selected by normalizing the attention map by its maximal value. Next, clipping the sum of the columns of the normalized attention map using a given threshold. A threshold of 0.3 was used, denoting all normalized values below it as 0 and above it as 1. The vast majority of values in the normalized attention map were below 0.1 due to the vast majority of elements being unsimilar and a large gap exists between the normalized values of 0.1 and 0.6 (see Appendix), thus the given threshold of 0.3 denotes a middle region dividing the two.

By investigating the nature of vector context similarities using the high attention map elements, a clear image emerges. Different heads place different emphasis on recognizing repeated tokens across different areas of the text. By contrast, others emphasize the recognition of various tokens and their connections (Fig. 6). Notably, the grammatical constraint of noticing similar tokens within the same sentence did not significantly affect the overall behavior of the heads or their recognition (Fig. 6b).

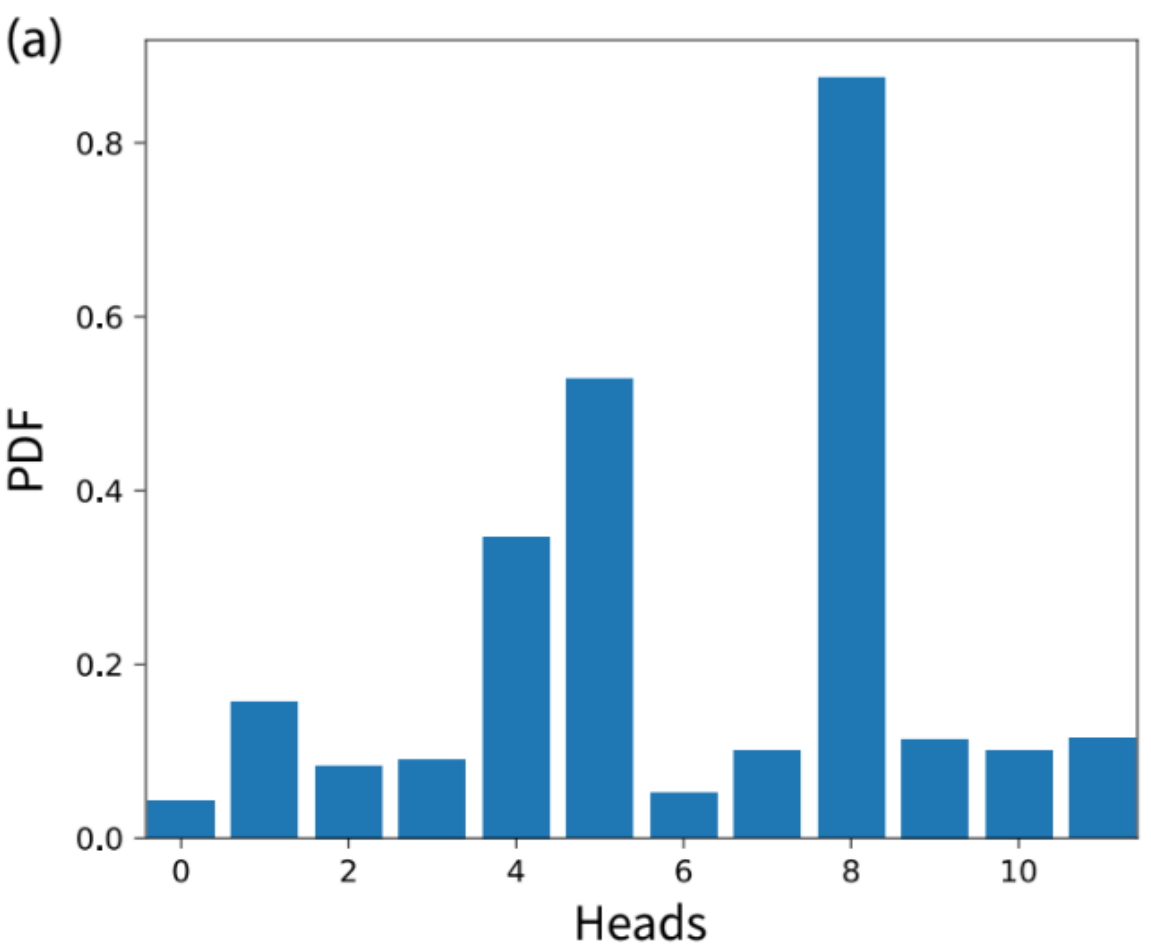

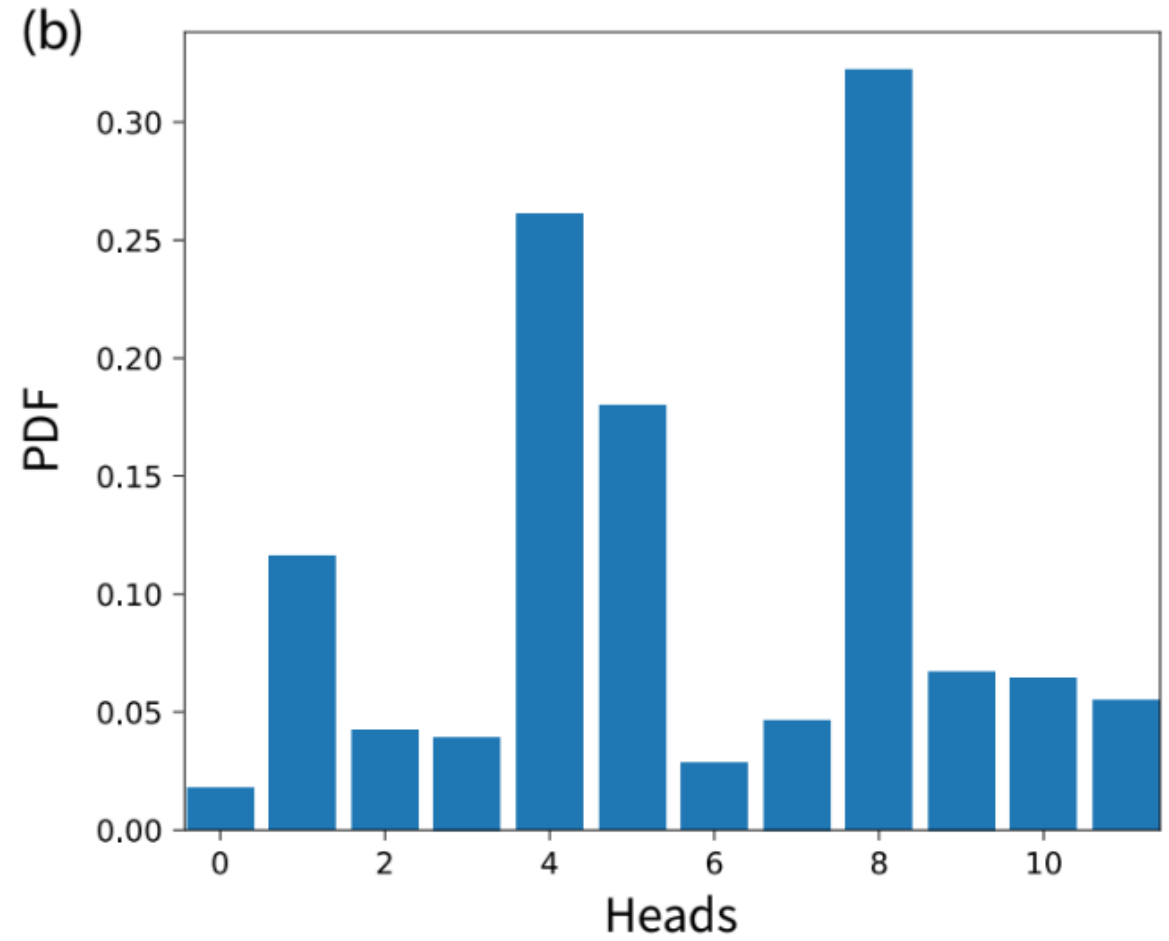

**Fig. 6.** (a) Probability distribution function of attention heads' recognition of repeating tokens within the text, that is, the probability that high context similarity elements are between equal token values. (b) Same as (a) but for recognizing repeating tokens in the same sentence.

Another aspect of the heads' behavior is the emphasis not on the token value but on the token's location and nature of the heads in relation to it. The distance, that is, the difference in location with respect to placement in the text, between token vector pairs with high context similarity can be extrapolated. Although all attention heads of Layer 11 show a sharper inclination to short-distance correlations, with the notable exception of head 8 (Fig. 7), certain heads are more inclined to locate on a marginally longer-range correlation. However, all heads have an average correlation distance below 50, except head 8, which recognizes identical tokens across the entire input text.

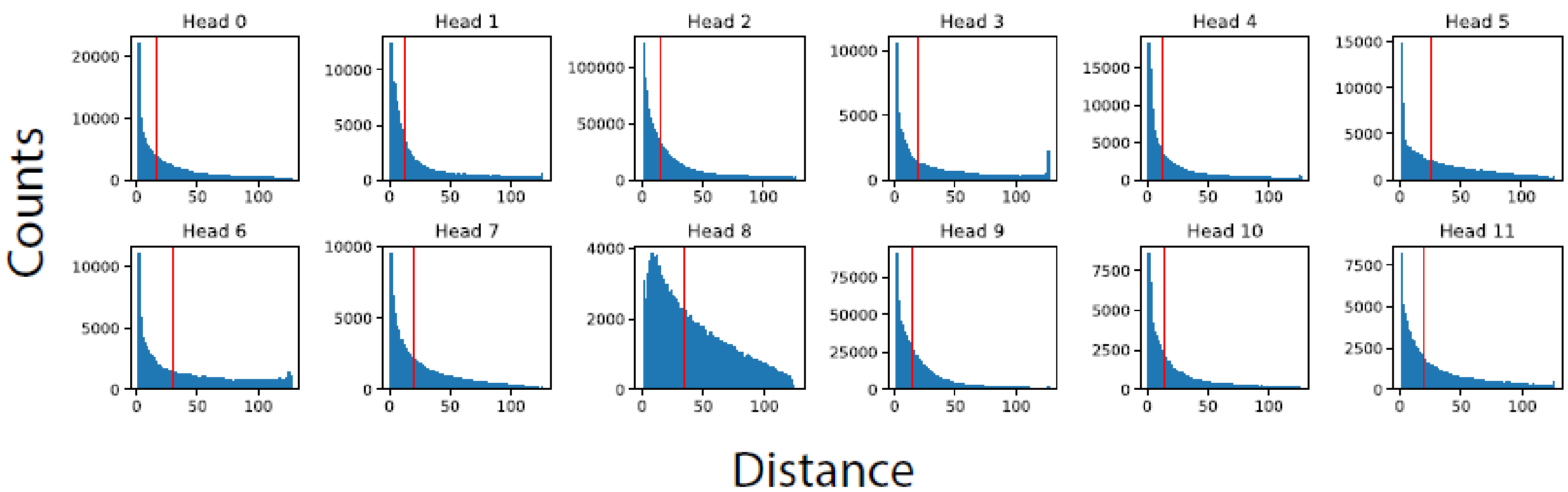

**Fig. 7.** Histogram of distances between token pairs with high context similarity between them per head over 20,000 text samples. The vertical red line in each panel stands for the median distance.

Short-distance correlation is also exemplified by the tendency of heads to create strong context similarities between tokens within the same sentence rather than across sentences. Because each sample of 128 tokens consists of around 3-4 sentences, the strong preference for heads to focus on the same sentence, exceeding 33%, was clearly an inclination derived from the learning process rather than a statistical phenomenon (Fig. 8).

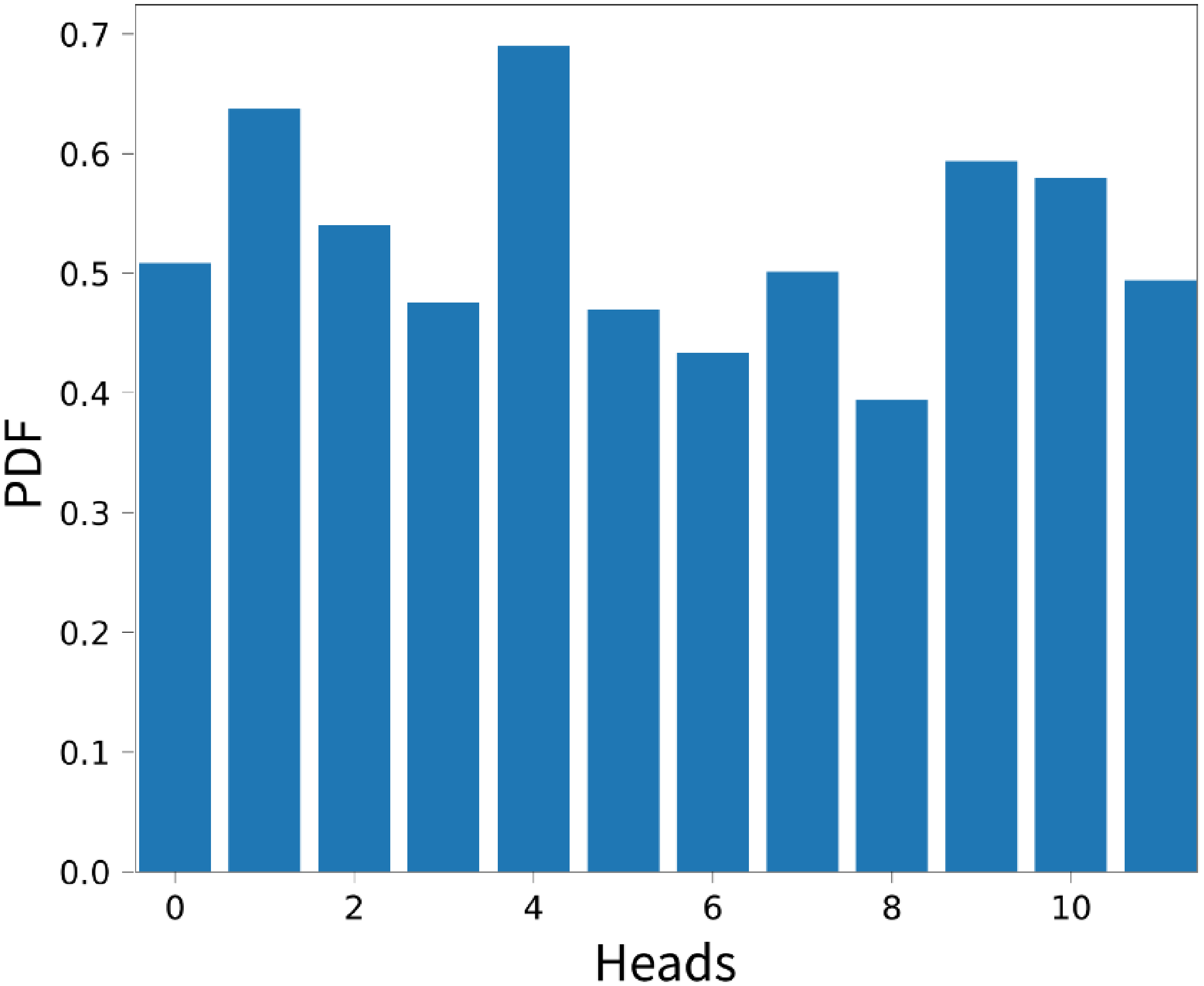

**Fig. 8.** Distribution probability function of high context similarity of two tokens within the same sentence per head.

The recognition of both similar and different tokens, as well as the recognition of short- and long-range correlations between tokens, is exemplified in Fig. 9, where exemplary heads have been chosen, and token pairs with high context similarity have been marked in the original text.

Head 7

| Token_i | Token_j | Index_i | Index_j |
|---|---|---|---|
| war | central | 22 | 42 |
| war | government | 22 | 43 |
| war | 1974 | 22 | 46 |
| war | act | 22 | 73 |
| civilian | central | 28 | 42 |
| civilian | government | 28 | 43 |
| civilian | 1974 | 28 | 46 |
| three | central | 41 | 42 |
| three | government | 41 | 43 |
| three | act | 41 | 73 |
| central | government | 42 | 43 |
| central | 1974 | 42 | 46 |
| central | act | 42 | 73 |
| government | 1974 | 43 | 46 |
| government | act | 43 | 73 |
| 1974 | territorial | 46 | 58 |
| 1974 | act | 46 | 73 |
| 1974 | minor | 46 | 77 |

Head 5

| Token_i | Token_j | Index_i | Index_j |
|---|---|---|---|
| post | war | 20 | 22 |
| major | major | 27 | 63 |
| saw | deployed | 54 | 61 |
| saw | switched | 54 | 85 |
| police | policing | 56 | 70 |
| police | police | 56 | 88 |
| deployed | switched | 61 | 85 |
| major | minor | 63 | 77 |
| policing | police | 70 | 88 |

Head 8

| Token_i | Token_j | Index_i | Index_j |
|---|---|---|---|
| airport | airport | 1 | 109 |
| in | the | 18 | 19 |
| in | in | 18 | 51 |
| airports | airport | 29 | 109 |
| airports | airport | 64 | 109 |
| from | to | 108 | 110 |
| airport | airport | 109 | 111 |
| airport | airport | 109 | 117 |

| 0: [CLS] | 16: airline | 32: responsibility | 48: when | 64: airports | 80: in | 96: , | 112: , |
|---|---|---|---|---|---|---|---|
| 1: airport | 17: services | 33: of | 49: the | 65: under | 81: size | 97: the | 113: leading |
| 2: policing | 18: in | 34: specialist | 50: rise | 66: the | 82: , | 98: funding | 114: to |
| 3: in | 19: the | 35: con | 51: in | 67: provisions | 83: they | 99: agreements | 115: disagreements |
| 4: the | 20: post | 36: ##sta | 52: international | 68: of | 84: too | 100: for | 116: between |
| 5: united | 21: - | 37: ##bular | 53: terrorism | 69: the | 85: switched | 101: the | 117: airport |
| 6: kingdom | 22: war | 38: ##ies | 54: saw | 70: policing | 86: to | 102: provision | 118: operators |
| 7: has | 23: period | 39: operated | 55: armed | 71: of | 87: armed | 103: of | 119: and |
| 8: taken | 24: . | 40: by | 56: police | 72: airports | 88: police | 104: such | 120: chief |
| 9: many | 25: policing | 41: three | 57: from | 73: act | 89: provided | 105: services | 121: constable |
| 10: forms | 26: at | 42: central | 58: territorial | 74: . | 90: by | 106: varied | 122: ##s |
| 11: since | 27: major | 43: government | 59: police | 75: as | 91: local | 107: wildly | 123: . |
| 12: the | 28: civilian | 44: departments | 60: forces | 76: more | 92: police | 108: from | 124: a |
| 13: rise | 29: airports | 45: until | 61: deployed | 77: minor | 93: forces | 109: airport | 125: new |
| 14: of | 30: was | 46: 1974 | 62: to | 78: airports | 94: . | 110: to | 126: regime |
| 15: scheduled | 31: the | 47: , | 63: major | 79: grew | 95: however | 111: airport | 127: [SEP] |

**Fig. 9.** Table showing exemplary heads and their recognition of similar and different tokens for a given input text sequence. Each color represents a different token value appearing in the text

(bottom) that composed  a high context similarity element appearing in one of the presented heads' (top left and top right) context similarity matrix.

The behavior of the heads was independent of the text presented. Accumulating areas of high context similarity across different heads reveals an agglomerative effect, with each head focusing on high context similarity in specific regions (Fig. 10).

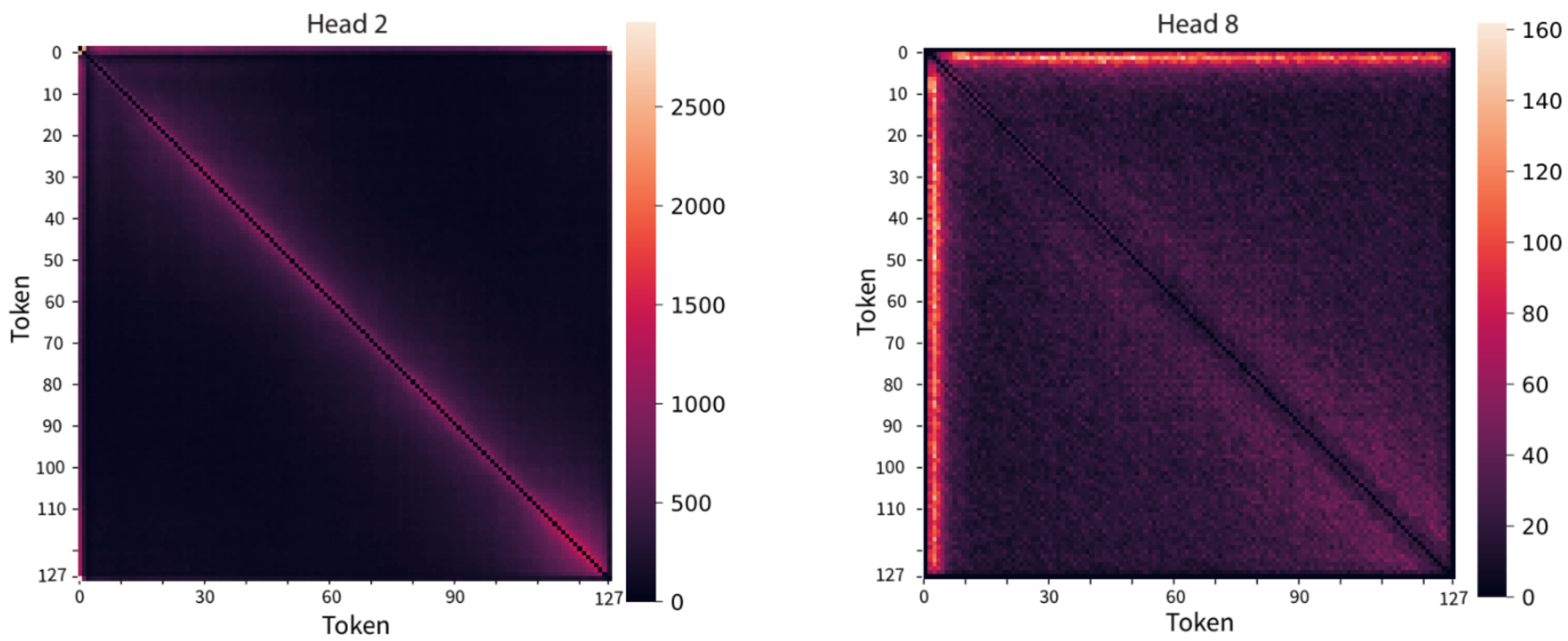

**Fig. 10.** Summation of strong context similarities over 20,000 text examples showing the most common locations of strong context similarities of Head 2 (left) and Head 8 (right).

Another phenomenon is the specialization of each head [8] over a unique token for each input example. For each head, the vast majority of its most influential context similarity elements were token pairs which included a single most common token with a relatively high probability (Fig. 11a). On average, each head demonstrated a tendency to specialize in a distinct token for each text sample rather than all the heads converging to a standard token (Fig. 11b). This means that the attention mechanism shows a clear tendency to assign a specific unique token to each head and have the head create strong context similarities of other tokens with it.

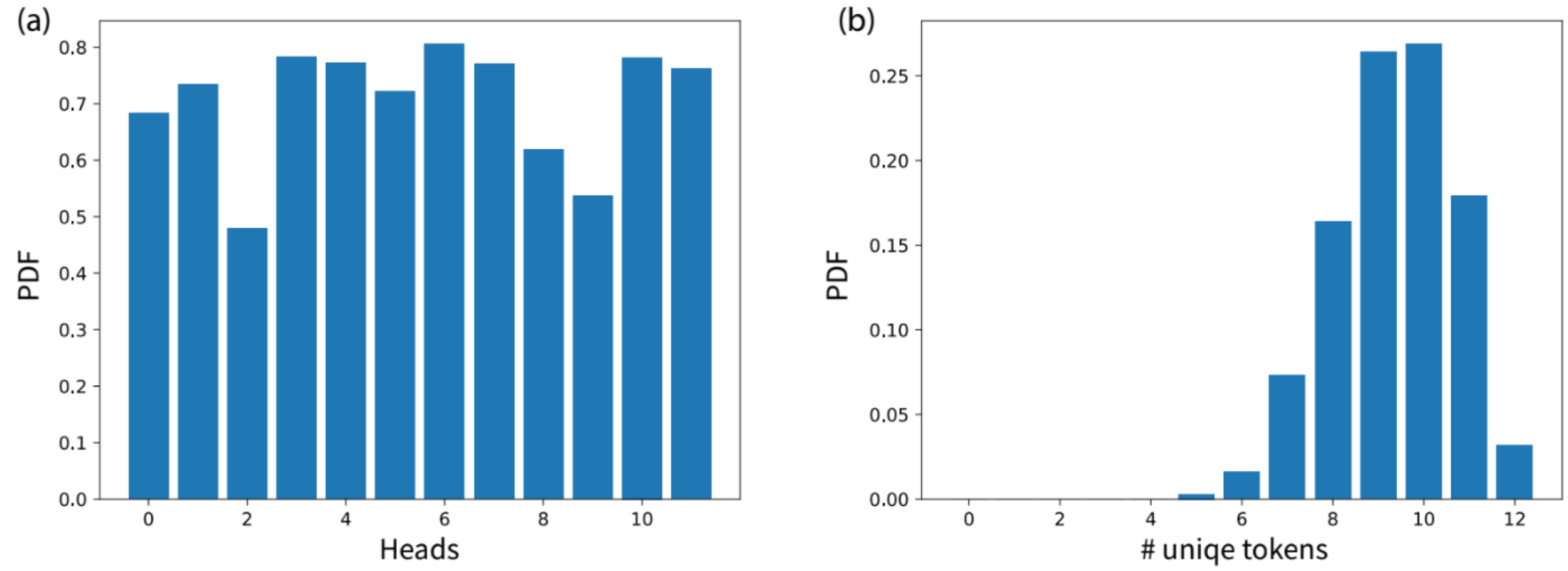

**Fig. 11.** (a) Probability distribution function of high context similarity elements that include their most common token per head. (b) Probability distribution function of the number of unique most common tokens (ranging from 0 to 12) across the layer's heads, averaged over 1,000 text examples.

## 4. Evolution through layers

Extrapolating the context-similarity matrices of the other layers paints a picture of the architecture's overall behavior. In the earlier layers, context similarity elements were more evenly distributed over the entire $128 \times 128$ token context similarity matrix (Fig. 12, left). As the layers advance, the context similarity becomes more concentrated along the diagonal, with elements that are highly similar along the diagonal and very low similarity elsewhere (Fig. 12, middle). As the architecture reaches the final layers, the context similarity spreads more, covering tokens bound together within the same sentences and a lower similarity outside the confines of the sentences (Fig. 12, right).

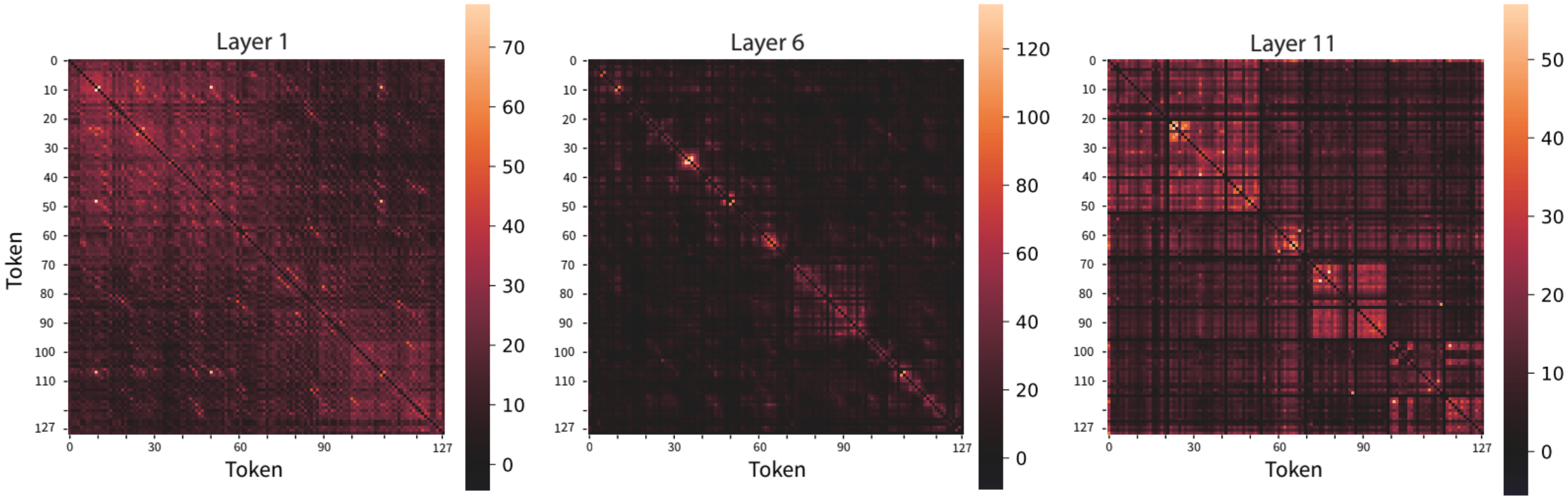

**Fig. 12.** Summed context similarity matrices of the 12 heads at layers 1 (left), 6 (middle), and 11 (right) of all the attention head outputs for a specified text input.

The distance distribution of high context similarity elements also changed with the advancement of the layers. Early layers are more spread out, creating almost random significance among the tokens regardless of their range. Extremely short-range correlations (relatively to the maximal distance 127) govern the middle layers almost entirely. The advanced layers' distance distributions are slightly longer, yet still exhibit short-range correlations (Fig. 13). This is consistent with the results of the context similarity map presented in Fig. 12 and with the nature of the learning process, which gradually learns the syntactic context as the layers advance (Fig. 4).

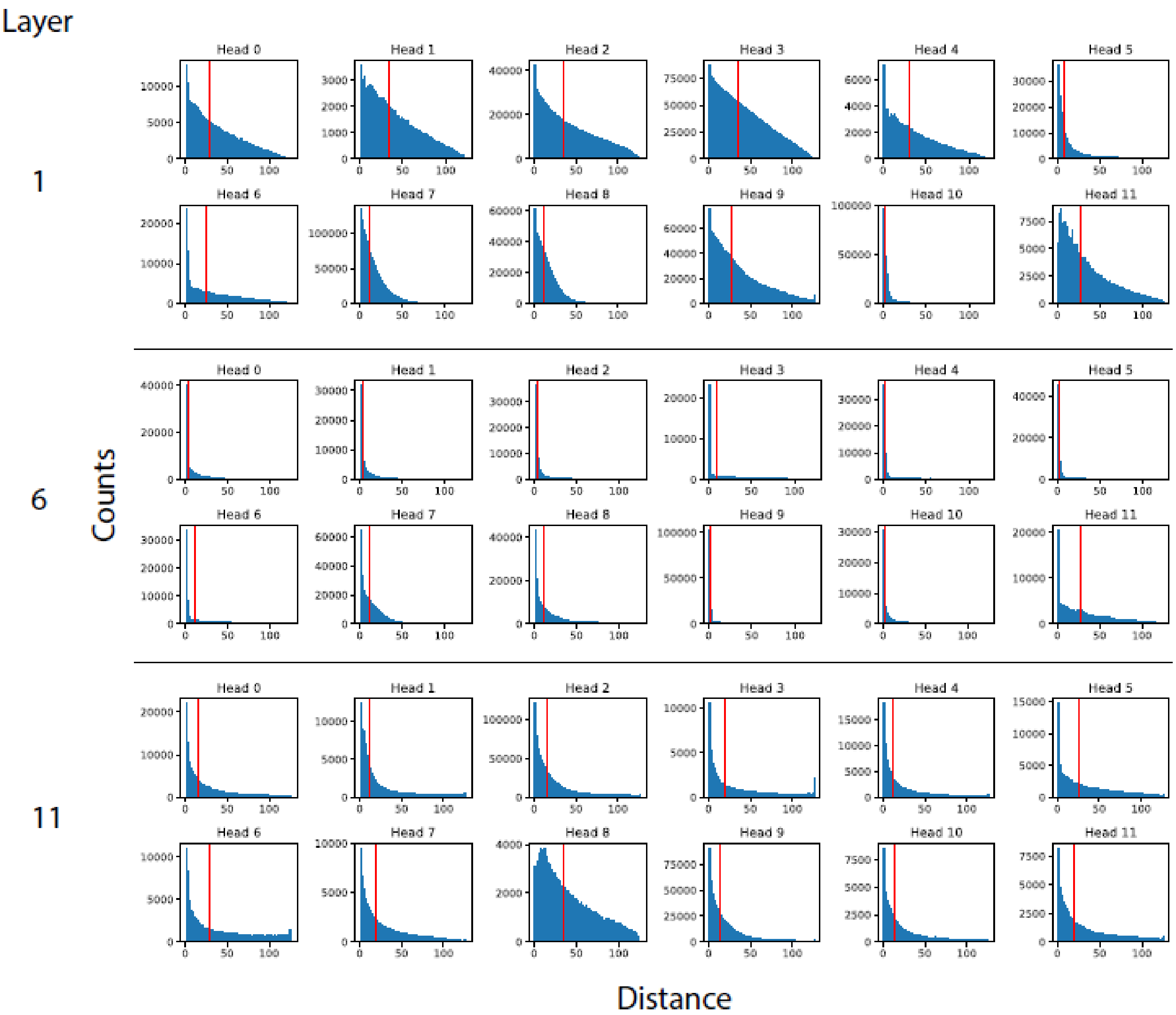

**Fig. 13.** Distance of significant context similarity elements of layers 1 (top), 6 (middle), and 11 (bottom), and median distance in each panel (vertical red line).

## 5. Discussion

Understanding the behavior of the vector space within transformer architectures can significantly deepen our understanding of NLP. Although attention maps provide an initial basic understanding of the architecture's mechanism, they may not fully capture how transformers convert contextual-linguistic information into a numerical representation. Notably, some researchers even argue that the attention map lacks any form of explainability [9, 12, 16]. The focus on the output of the attention heads rather than the attention maps of each head presents a more fundamental view of how the vectors corresponding to each token in the text alternate and how they are refined as the layers progress, as well as the interplay between them within the given layers. The attention map is also incapable of explaining macroscopic phenomena governing learning, as a universal law explaining where areas of high attention form and why they form there is lacking [12, 16].

The vector space emerging from the embedding layer represents a rudimentary transformation of tokens into a vector space focused solely on lexical meaning [17] which is then altered by the transformers and concentrates on regions of syntactical relevance. As the layers progress, the context similarities converge from the overall averaged similarity across all tokens to a centralized similarity along the diagonal, then expand to syntactical separators. This is also exemplified by the token-distance distribution of high-context similarity elements within the context similarity matrix. The emergence of such behavior, examined through the context similarity map, illustrates the explainable capabilities of the context similarity matrix and how effectively it exemplifies the information processing performed by the attention architecture.

The behavior of the emergent vector space raises two crucial questions. First, how can contextual comprehension of long-range correlations be achieved? Although the model is capable of capturing medium-range correlations within sentences, it has not yet demonstrated the ability to handle long-range contextual relationships across larger texts. Addressing this issue can significantly advance the field of NLP. Long-range interactions are not as common in physical systems, which typically act on short-range interactions among their individual elements, such as in spin-glass models. This indicates that solutions may lie in architectures capable of performing direct interactions among information presented over long ranges, such as biological learning systems, which excel at understanding language. Second, the question arises as to whether the vector space, in its current form, is in its optimal state. Similar vectors share similar contextual–semantic meanings; however, the vector space created by the transformer architecture contains noise elements, evident through high-context similarity elements with no direct linguistic comprehension (Fig. 5). Noise mitigation remains an ongoing challenge in learning systems.

Different systems have developed methods to reduce noise using the learning process itself [18, 19]. A better understanding of the signal and noise elements propagated through the architecture can lead to more efficient pruning techniques or dilution strategies, thereby significantly reducing latency and memory usage [20, 21]. However, although mitigating noise in an existing system is attainable, this raises the question of whether the presented attention-mechanism architecture is optimal. Previous studies have questioned the necessity of large embedding dimensions [22, 23] and concerns have been raised about the complexity and costs of these large systems [24]. Developing more efficient and smarter alternatives to advanced attention techniques that support long-range interactions is necessary for improved contextual understanding.

The specialization of heads across different tokens indicates that the architecture is attempting to recognize distinct components of the sentence structure. This tendency is a replication of a phenomenon also observed in computer vision tasks [6, 25, 26] where filters and attention heads specialize in specific regions and collectively accumulate all the information in the input. Notably, although a clear classification task is presented in computer vision tasks [27, 28] NLP tasks do not always present such well-defined objectives. This raises the question of how transformer architectures perform in NLP classification tasks [1, 29, 30] *whether the current learning methods are optimal for these tasks* [17].

Finally, this study proposes an initial, quantifiable language to assess the learning performance of transformer architectures. Nevertheless, the need for a quantifiable language for Signal-to-Noise ratio (SNR) persists. Although the presentation of a context similarity matrix can serve as a more quantifiable language, a more comprehensive understanding of the SNR governing this architecture can shed significant light on how transformers learn. Furthermore, a more comprehensive analysis of the context similarity matrices of other architectures is required in order to assess its universality, as other features have seen to be universal across other deep learning tasks and models[17, 18, 31, 32].It will also be beneficial to compare and contrast their behavior and that of encoder-decoder architectures, such as translation systems [33, 34]. , it was observed that as the transformer blocks progress, they become more capable of filling masked words in sentences in an attempt to lower the language entropy of the system [17]. However, translating this tokenized accuracy into a vectorized signal and noise requires further extensive research.

**Appendix**

Distance measurement: For Figs. 7 and 11, the distance between two tokens is defined as the absolute difference between their positions in the sequence. For each highly context-similar token pair with indices $(i, j)$ , the distance is given by $|i - j|$.

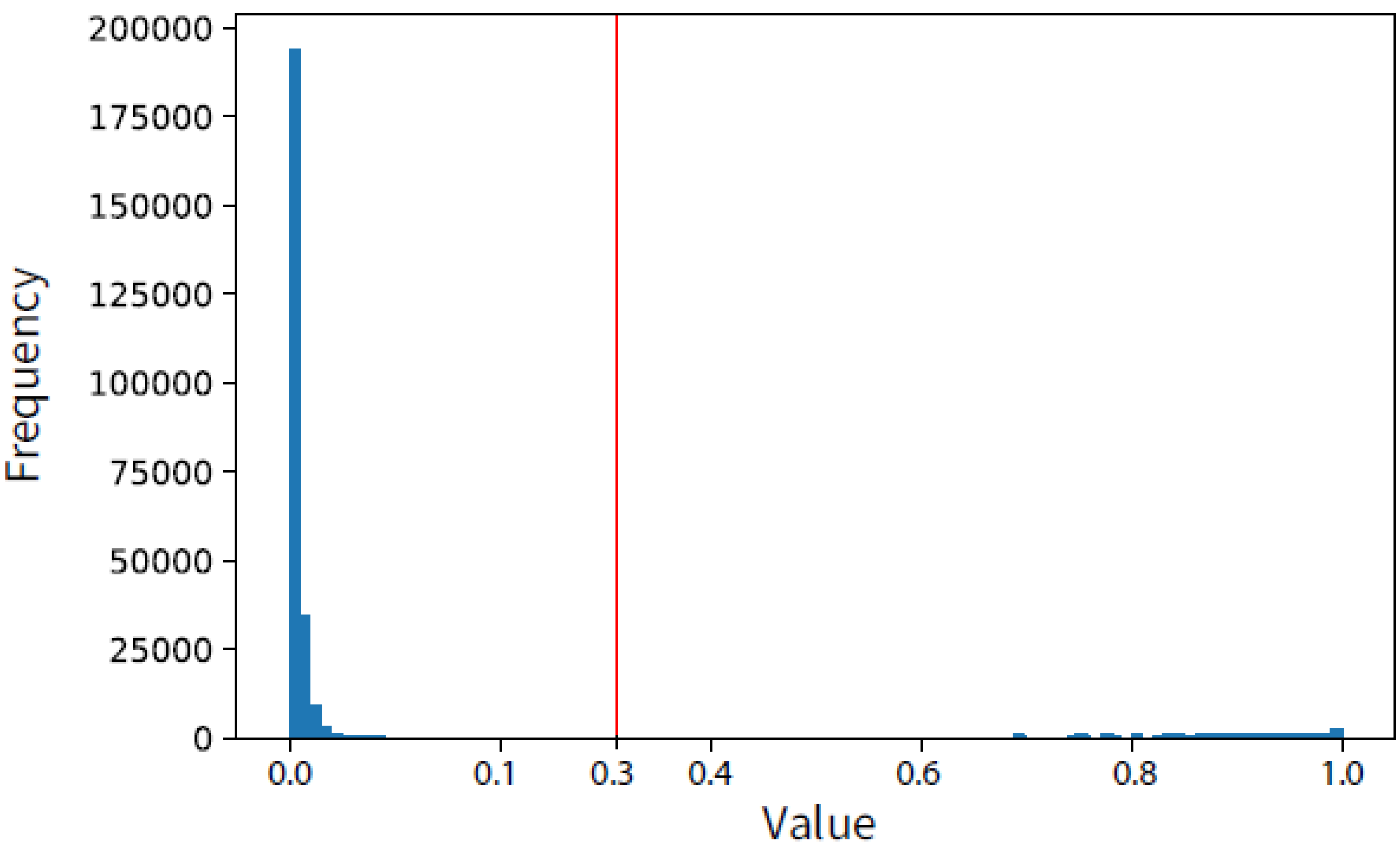

**Fig. A1.** Columns sum distribution, taken from 2000 sequences of 128 tokens of head 11 in layer 11, indicates a large gap around a threshold 0.3 (vertical red line). The attention map is first normalized by its maximal value.

*Hardware and software:* We used Google Colab Pro and its available GPUs. We used Pytorch for all the programming processes. The HuggingFace pre-trained BERT-12 architecture was used for all tests and results.